\documentclass{article}
\usepackage{ijcai17}
\usepackage{times}

\usepackage[english]{babel}
\usepackage[utf8x]{inputenc}
\usepackage{amsmath,amssymb}
\usepackage{graphicx}
\usepackage[colorinlistoftodos]{todonotes}
\usepackage{bm}
\usepackage{algorithm}
\usepackage{algpseudocode}
\usepackage{multirow}
\usepackage{caption}
\usepackage{subcaption}

\title{Options Discovery with Budgeted Reinforcement Learning}
\author{
  Aurélia Léon and Ludovic Denoyer\\
  Sorbonne Universités,\\ 
  UPMC Univ Paris 06, UMR 7606, LIP6, F-75005, Paris, France\\
  \{aurelia.leon,ludovic.denoyer\}@lip6.fr
}

\begin{document}

\maketitle

\begin{abstract}
In hierarchical reinforcement learning, the framework of options models sub-policies over a set of primitive actions. In this paper, we address the problem of discovering and learning options from scratch. Inspired by recent works in cognitive science, our approach is based on a new budgeted learning approach in which options naturally arise as a way to minimize the \textit{cognitive effort} of the agent. In our case, this effort corresponds to the amount of information acquired by the agent at each time step. We propose the \textit{Budgeted Option Neural Network} model (BONN), a hierarchical recurrent neural network architecture that learns latent options as continuous vectors. With respect to existing approaches, BONN does not need to explicitly predefine sub-goals nor to \textit{a priori} define the number of possible options. We evaluate this model on different classical RL problems showing the quality of the resulting learned policy.
\end{abstract}

\section{Introduction}


Works in cognitive science have long emphasized that human or animal behavior can be seen as a hierarchical process, in which solving a task amounts to sequentially solving sub-tasks  \cite{hie}. Examples of those in a simple maze environment are \textit{go to the door} or \textit{go to the end of the corridor}; these sub-tasks themselves correspond to sequences of primitive actions or other sub-tasks.

In the computer science domain, these works have led to the \textit{hierarchical reinforcement learning} paradigm \cite{dayan1993feudal,dietterich1998maxq,parr1998reinforcement}, motivated by the idea that it makes the discovery of complex skills easier.  One line of approaches consists in modeling sub-tasks through \textit{options} \cite{sutton1999between}, giving rise to two main research questions: (a) choosing the best suited option and then (b) selecting actions to apply in the environment based on the chosen option. These two challenges are respectively mediated using a high-level and a low-level controllers. 

In the Reinforcement Learning (RL) literature, different models have been proposed to solve these questions. But the additional question of \textit{discovering options} is rarely addressed: the hierarchical structure in the majority of existing models has to be manually constrained, e.g. by predefining possible sub-goals \cite{kulkarni2016hierarchical}. Learning automatic task decomposition, without supervision, remains an open challenge in the field of RL. The difficulty of discovering hierarchical policies is emphasized by the fact that it is not well understood \textit{how} they emerge in behaviors.



Recent studies in neurosciences suggest that a hierarchy can emerge from \textit{habits} \cite{dezfouli2013actions}. Indeed, \cite{keramati2011speed} and \cite{dezfouli2012habits} distinguish between \textit{goal-directed} and \textit{habitual} action control; the latter not using any information from the environment.  The underlying idea is that the agent first uses goal-directed control, and then progressively switches to habitual action control since goal-directed control requires a higher \textit{cognitive cost} to process the acquired information\footnote{In that case, the cognitive cost is the time the animal spends to decide which action to take.}. \hfill~\linebreak

\begin{figure}[t]
\begin{subfigure}[t]{0.23\textwidth} 
\includegraphics[width=0.95\linewidth]{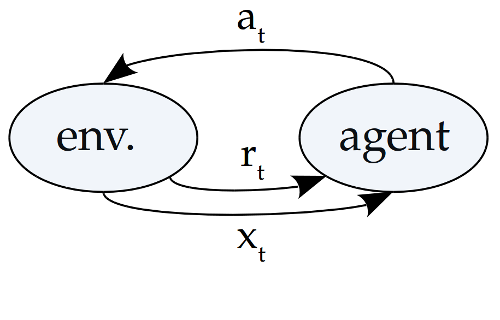}
\caption{}
\end{subfigure}
\begin{subfigure}[t]{0.23\textwidth}
\includegraphics[width=0.9\linewidth]{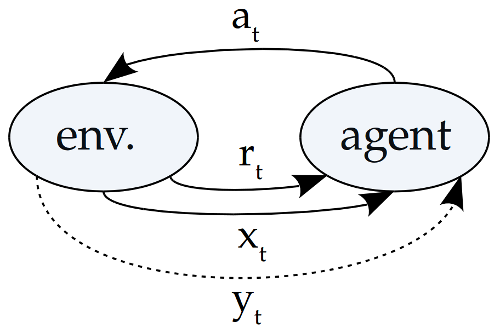}
\caption{}
 \end{subfigure}
\caption{(a) A typical RL setting where the agent receives a reward $r_t$ and an observation $x_t$ from the environment at each time step, then executes an action $a_t$. (b) The setting used for the BONN model, where the agent can ask for an additional information $y_t$. }\vspace{-0.3cm} 
\label{fig/RL_y}
\centering
\end{figure}

Based on this idea, we propose the BONN architecture (Budgeted Options Neural Network). In this framework, the agent has access to different amounts of information coming from the environment. More precisely, we consider that at each time step $t$, the agent can use a basic observation denoted $x_t$ used by the \textit{low-level} controller as in classical RL problems. In addition, it can also choose to acquire a supplementary observation $y_t$, as illustrated in Fig. \ref{fig/RL_y}. This extra observation will provide more information about the current state of the system and will be used by the \textit{high-level} controller, but at a higher ``cognitive'' cost. On top of this setting, we assume that options will naturally emerge as \textbf{a way to reduce the overall cognitive effort generated by a policy}. In other words, by constraining the amount of high-level information used by our system (i.e the $y_t$), BONN will learn a hierarchical structure where the resulting policy is a sequence of low-cost sub-policies. In that setting, we show that the structure of the resulting policy can be seen as a sequence of intrinsic options \cite{gregor2016variational}, i.e. vectors in latent space.

The contributions of the paper are threefold:
\begin{itemize}
  \item We propose a new assumption about options discovery, arguing that it is a consequence of learning a trade-off between policy efficiency and cognitive effort. The hierarchical structure emerges as a way to reduce the overall cost.
\item We define a new model called BONN that implements this idea as a hierarchical recurrent neural network for which, at each time step, two observations are available at two different prices. This model is learned using a policy gradient method over a budgeted objective.
\item We propose different sets of experiments in multiple settings showing the ability of our approach to discover relevant options.
\end{itemize}
 
The paper is organized as follows: related works are presented in Section \ref{rw}. We define the background for RL and for recurrent policy gradient methods in Section \ref{sec2}. We introduce the BONN model and the learning algorithm for the budgeted problem in Section \ref{sec3}. Experiments are presented and discussed in Section \ref{exp}. At last, Section 6 concludes and opens perspectives.

\section{Related Work}
\label{rw}

The closest architecture to BONN is the Hierarchical Multiscale Recurrent Neural Network \cite{chung2016hierarchical} that discovers hierarchical structures in sequences. It uses a binary boundary detector learned with a straight-through estimator, similar to the acquisition model (see Section \ref{sec32}) of BONN.i

More generally, \textbf{Hierarchical Reinforcement Learning} \cite{dayan1993feudal,dietterich1998maxq,parr1998reinforcement} has been the surge of many different works during the last decade since it is deemed as one solution to solve long-range planning tasks and to allow the transfer of knowledge between tasks. Many different models assume that subtasks are \textit{a priori} known, e.g., the MAXQ method in \cite{dietterich1998maxq}.  The concept of \textbf{option} is introduced by \cite{sutton1999between}. In this architecture, each option consists of an initiation set, its own policy (over primitive actions or other options), and a termination function which defines the probability of ending the option given a certain state. The concept of options is at the core of many recent articles. For example, in \cite{kulkarni2016hierarchical}, the authors propose an extension of the Deep Q-Learning framework to integrate hierarchical value functions using intrinsic motivation to learn the option policies.  But in these different models, the options have to be manually chosen \textit{a priori} and are not discovered during the learning process. 
Still in the options framework, \cite{danielprobabilistic} and \cite{bacon2015option} discover options (both internal policies and the policy over options) without supervision, using respectively the Expectation Maximization algorithm and the option-critic architecture. Our contribution differs from these last two in that the BONN model does not have a fixed discrete number of options and rather uses an ``intrinsic option'' represented by a latent vector. Moreover, we clearly state how options arise by finding a good trade-off between efficiency and cognitive effort.

Close to our work, the concept of cognitive effort was introduced in \cite{bacondeliberate} (but with discrete options), while  intrinsic options, i.e. options as latent vectors, where used in \cite{gregor2016variational}.


At last, some articles propose hierarchical policies based on \textbf{different levels of observations}. A first category of models is those that use open-loop policies i.e. do not use observation from the environment at every time step. \cite{hansen1996reinforcement} propose a model that mixes open-loop and closed-loop control while considering that sensing incurs a cost. Some models focus on the problem of learning \textbf{macro-actions} \cite{hauskrecht1998hierarchical,mnih2016strategic}: in that case, a given state is mapped to a sequence of actions. 
Another category of models divides the state space into several components. For instance, the Abstract Hidden Markov Model \cite{ahmm} is based on discrete options defined on each space region. 
\cite{heess2016learning} use a low-level controller that has only access to the proprioceptive information, and a high-level controller has access to all observations. 
\cite{florensa2016stochastic} use a similar idea of factoring the state space into two components, and learn a stochastic neural network for the high-level controller. 
The \textit{blind setting} of the BONN model, described in Section \ref{section/blind}, is similar to (stochastic) macro-actions, but open-loop policies are rather limited in complex environments. The general BONN architecture is more comparable to works using two different observations, however, those models do not learn \textit{when} to use the high-level controller.

\section{Background}
\label{sec2}

\subsection{ (PO-) Markov Decision Processes and Reinforcement Learning}

Let us denote a Markov Decision Process (MDP) as a set of states $\mathcal{S}$, a discrete set of possible actions $\mathcal{A}$, a transition distribution $P(s_{t+1}|s_t,a_t)$ and a reward function $r(s_t,a_t) \in \mathbb{R}^+$. We consider that each state $s_t$ is associated with an observation $x_t \in \mathbb{R}^n$, and that $x_t$ is a partial view of $s_t$ (i.e POMDP), $n$ being the size of the observation space. Moreover, we denote $P_I$ the probability distribution over the possible initial states of the MDP. 

Given a trajectory $x_0,a_0,x_1,a_1,....,x_t$, a policy is defined by a probability distribution such that $\pi(x_0,a_0,x_1,a_1,....,x_t,a)=P(a|x_0,a_0,x_1,a_1,....,x_t)$ which is the probability of each possible action $a$ at time $t$, knowing the history of the agent. 

\subsection{Learning with Recurrent Policy Gradient}

Let us denote $\gamma \in ]0,1]$ the discount factor, and $R_t=\sum\limits_{k=t}^{T-1} \gamma^{k-t} r(s_k,a_k)$ the discounted sum of rewards (or discount return) at time $t$, corresponding to the trajectory $(s_t,a_t,s_{t+1},a_{t+1},....,s_T)$ with $T$ the size of the trajectories sampled by the policy \footnote{We describe finite-horizon problems where $T$ is the size of the horizon and $\gamma \leq 1$, but the approach can also be applied to infinite horizon problems with discount factor $\gamma<1$}. Note that $R_0=\sum\limits_{t=0}^{T-1} \gamma^{t} r(s_t,a_t)$ corresponds to the classical discount return.

We can define the reinforcement learning problem as the optimization problem such that the optimal policy $\pi^*$ is computed by maximizing the expected discounted return $J(\pi)$:
\begin{equation}
J(\pi)=\mathbb{E}_{s_0 \approx P_I;a_0,....,a_{T-1} \approx \pi}\left[R_0\right]
\end{equation}
where $s_0$ is sampled following $P_I$ and the actions are sampled based on $\pi$.

Different learning algorithms aim at maximizing $J(\pi)$. In the case of policy gradient techniques, if we consider that, for sake of simplicity, $\pi$ also denotes the set of parameters of the policy, the gradient of the objective can be approximated with:
\begin{equation}
\nabla_\pi J(\pi) \approx \frac{1}{M} \sum\limits_{m=1}^M \sum\limits_{t=0}^{T-1}  \nabla_\pi \log \pi(a_t|x_0,a_0,...,x_t)  \left(R_t - b_t \right) 
\end{equation}
where $M$ is the number of sampled trajectories used for approximating the gradient using Monte Carlo sampling techniques, $b_t$ is a variance reduction term at time $t$ estimated during learning, and we consider that future actions do not depend on past rewards (see \cite{rpg} for details on recurrent policy gradients).

\section{Budgeted Option Neural Network}
\label{sec3}

\subsection{The BONN Architecture}

\label{sec32}

In a typical (PO-)MDP setting, the agent uses an observation $x_t$ from the environment at every time step. In contrast, in the BONN architecture, the agent always uses a \textit{low-level} observation $x_t$, but can also choose to acquire a \textit{high-level} observation $y_t$ that will provide a more relevant information, as illustrated in Figure \ref{fig/RL_y}. 
This situation corresponds to many practical cases: for example a robot that acquires information through its camera ($x_t$) can sometimes decide to make a complete scan of the room ($y_t$); a user driving a car (using $x_t$) can decide to consult its map or GPS ($y_t$); a virtual agent taking decisions in a virtual world (based on $x_t$) can ask instructions from a human ($y_t$), etc. 
Note that a particular case (called \textit{blind setting}) is when $x_t$ is empty, and $y_t$ is the classical observation over the environment. In that case, the agent will basically decide whether it wants to use the current observation, as in the \textit{goal directed} vs \textit{habits} paradigm \cite{keramati2011speed,dezfouli2012habits}.
\hfill~\linebreak

\begin{figure}
\includegraphics[width=0.99\linewidth]{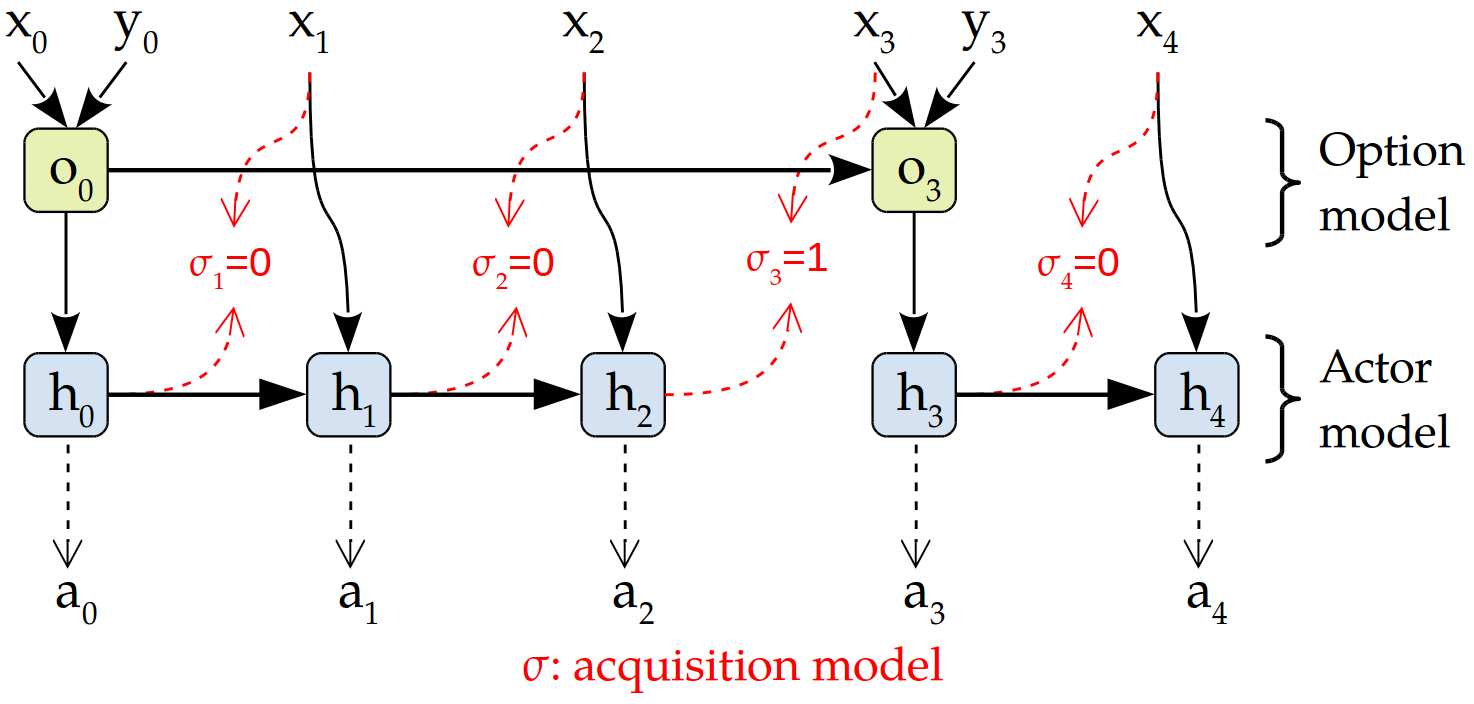}
\caption{The BONN Architecture. Arrows correspond to dependencies, dashed arrows correspond to sampled values. Note that in this example, $\sigma_3=1$ so the model decides to acquire $y_3$ and computes a new option $o_3$,. When $\sigma_t=0$ (for $t\in \{1,2,4\}$ in this example) the model does not use $y_t$ and keeps the same option.}
\label{fig/archi}
\end{figure}

The structure of BONN is close to a hierarchical recurrent neural network with two hidden states, $o_t$ and $h_t$, and is composed of three components. The \textbf{acquisition model} aims at choosing whether observation $y_{t}$ has to be acquired or not. 
If the agent decides to acquire $y_t$, the \textbf{option model} uses both observations $x_t$ and $y_t$ to compute a new option denoted $o_t$ as a vector in an option latent space. The \textbf{actor model} updates the state of the actor $h_t$ and aims at choosing which action $a_t$ to perform.


We now formally describe these three components. A schema of the BONN architecture is provided in Fig. \ref{fig/archi}, and the complete inference procedure is given in Algorithm \ref{alg}. Note that for sake of simplicity, we consider that the last chosen action $a_{t-1}$ is included in the low-level observation $x_t$, as often done in reinforcement learning, and we avoid to explicitly write the last chosen action in all equations. Relevant representations of $x_t$ and $y_t$ are learned through neural network -- linear models in our case. In the following, the notations $x_t$ and $y_t$ directly denote these representations, used as inputs in the BONN architecture.

\begin{algorithm}[t]
\caption{The pseudo code of the inference algorithm for the BONN model.}\label{alg}
\begin{algorithmic}[1]
\Procedure{Inference}{$s_0$}\Comment{$s_0$ is the initial state}
\State initialize $o_{-1}$ and $h_{-1}$ 
\For{$t=0$ to $T$}
\State Acquire $x_t$
\State \textit{acquisition model: } Draw $\sigma_t \in \{0,1\}$
\If{$\sigma_t==1$}
	\State \textit{option level: }Acquire $y_{t}$ and update the option state $o_t$
    \State \textit{actor level: } Initialize the actor state $h_t$
\Else
	\State \textit{actor level: } Update the actor state $h_t$
\EndIf
\State \textit{actor level: } Choose the action $a_t$ w.r.t $h_{t}$
\State \textbf{Execute the chosen action}
\EndFor
\EndProcedure
\end{algorithmic}
\end{algorithm}

\paragraph{Acquisition Model: }

The acquisition model aims at deciding whether a new high-level observation $y_t$ needs to be acquired. It draws $\sigma_t \in \{0,1\}$ according to a Bernoulli distribution with $P(\sigma_{t}=1)= sigmoid(f_{acq}(h_{t-1},x_{t}))$. 
If $\sigma_t=1$, the agent will use $y_t$ to compute a new option (see next paragraph), otherwise it will only use $x_t$ to decide which action to apply\footnote{In our experiments, $f_{acq}$ is a simple linear model following the concatenation of $h_{t-1}$ and $x_t$}.


\paragraph{Option Model:} If $\sigma_t=1$, the option model computes a new option state following $o_t = gru_{opt}(x_t,y_t,o_{last})$ in which $o_{last}$ is the lastly computed option before time step $t$. $gru_{opt}$ represents a GRU cell \cite{cho2014learning}.

\paragraph{Actor Model: } The actor model updates the actor state $h_t$ and computes the next action $a_t$. The update of $h_t$ depends on $\sigma$ and the availability of the high-level observation $y_t$:
\begin{itemize}
\item if $\sigma_t=0$: $h_t = gru_{act}(x_t,h_{t-1}) $
\item if $\sigma_t=1$: $h_t = o_t$
\end{itemize}
The next action $a_t$ is then drawn from the distribution $softmax(f_{act}(h_t))$\footnote{$gru_{act}$ represents a GRU cell while $f_{act}$ is in our experiments a simple perceptron}.





\paragraph{Options and  BONN: } When a new high-level observation is acquired i.e $\sigma_t=1$, the option state $o_t$ is updated based on the current observations $x_{t}$ and $y_{t}$. Then, the policy will behave as a classical recurrent policy until a next high-level observation is acquired. In other words, when acquiring a new high-level observation, a new sub-policy is chosen depending on $x_{t}$ and $y_t$ (and eventually the previous option). The option state $o_t$ can be seen as a latent vector representing the \textit{option} chosen at time $t$, while $\sigma_t$ represents what is usually called the \textit{termination function} in option settings. In BONN, since the option is chosen directly according to the state of the environment (more precisely on the observations of the agent $x_t$ and $y_t$), there is no need to have an explicit initiation set defining the states where an option can begin.

\subsection{Budgeted Learning for Options Discovery}
\label{sec33}


Inspired by cognitive sciences \cite{kool2014labor}, BONN considers that discovering options aims at reducing the \textit{cognitive effort} of the agent. In our case, the cognitive effort is measured by the amount of high-level observations $y_t$ acquired by the model to solve the task, and thus by the amount of options vectors $o_t$ computed by the model over a complete episode. By constraining our model to find a good trade-off between policy efficiency and the number of high-level observations acquired, BONN discovers when this extra information is essential and has to be acquired, and thus when to start new sub-policies.



Let us denote $C=\sum\limits_{t=0}^{T-1} \sigma_t$ the acquisition cost for a particular episode. We propose to integrate the acquisition cost $C$ (or cognitive effort) in the learning objective, relying on the \textit{budgeted learning} paradigm already explored in different RL-based applications \cite{contardo2016recurrent,dulac2012sequential}. We define an augmented immediate reward $r^*$ that includes the generated cost:
\begin{equation}
r^*(s_t,a_t,\sigma_t)=r(s_t,a_t)-\lambda \sigma_t 
\end{equation}
where $\lambda$ controls the trade-off between the policy efficiency and the cognitive charge. The associated discounted return is denoted $R_t^*$, so that $R_0^*$ will be used as the new objective to maximize
, resulting in the following policy gradient update rule: 
\begin{multline}
\pi \leftarrow \pi - \gamma \sum\limits_{t=0}^{T-1} ( \nabla_\pi \log P(a_t|h_t)  \\
+ \nabla_\pi \log P(\sigma_t|h_{t-1},x_t) ) \left( R_t^* - b^*_t \right)
\end{multline}
where $\gamma$ is the learning rate. Note that this rule now updates both the probabilities of the chosen actions $a_t$ and the probabilities of the $\sigma_t$ that can be seen as \textit{internal actions} and that decide if a new option has to be computed or not. $b^*_t$ is the new resulting variance reduction term as defined in \cite{rpg}.  

\subsection{Discovering a discrete set of options}
\label{discr}
In the previous sections, we considered that the option $o_t$ generated by the option model is a vector in a latent space. This is slightly different than the classical option definition which usually considers that an agent has a given ''catalog'' of possible sub-routines i.e the set of options is a finite discrete set. We propose here a variant of the model where the model learns a finite discrete set of options. 

Let us denote $K$ the (manually-fixed) number of options one wants to discover. Each option will be associated with a (learned) embedding denoted $o^k$. The \textit{option model} will store the different possible options and choose which one to use each time an option is required. In that case, the option model will be considered as a stochastic model able to sample one option index denoted $i_t$ in $\{1,2,...,K\}$ by using a multinomial distribution on top of a softmax computation. In that case, as the option model computes some stochastic choices, the policy gradient update rule will integrate these additional internal actions with:

\begin{multline}
\pi \leftarrow \pi - \gamma \sum\limits_{t=0}^{T-1} ( \nabla \log P(a_t|z_t) 
+ \nabla \log P(\sigma_t|z_{t-1},a_{t-1},x_t) \\
+  \nabla \log P(i_t|y_t) ) \left( R_t^*  - b_t \right)
\end{multline}
By considering that $P(i_t|y_t)$ is computed based on a softmax over a scoring function $P(i_t|y_t) \approx \ell(o_{i_t},y_t)$ where $\ell$ is a differentiable function, the learning will update both the $\ell$ function and the options embedding $o_k$. 

\section{Experiments}
\label{exp}



\subsection{Experimental setting}

For all experiments, we used the ADAM optimizer \cite{kingma2014adam} with gradient clipping\footnote{An open-source version of the model is available at https://github.com/aureliale/BONN-model.}
The learning rates were optimized by grid-search.

Observations $x_t$ and $y_t$ are represented through linear models with a hidden layer of sizes $N_x$ and $N_y$ respectively, followed by an activation function $relu$, and the GRU cells are of size $N_{gru}$ (dependent on the environment) \cite{cho2014learning}.

\subsection{Blind Setting}
\label{section/blind}

Given a POMDP (or MDP), the easiest way to design the two observations $x_t$ and $y_t$ needed for the BONN model is to consider that $x_t$ is the empty observation and $y_t$ is the usual observation coming from the environment. This case is similar to the ``\textit{goal-directed} vs \textit{habit action control}'' paradigm and corresponds to a case in which the agent chooses either to acquire or not the observation. It also corresponds to a (stochastic) macro-actions framework: the agent chooses a sequence of actions for each observation.

\begin{table}[t]
\begin{center}
\scriptsize{
\begin{tabular}{c|c|cc|cc}
& &\multicolumn{2}{c|}{$\epsilon=0$}&\multicolumn{2}{c}{$\epsilon=0.25$} \\
\hline
& &R&\%obs  & R&\%obs \\
\hline
\multirow{2}{*}{Cartpole}  &R-PG & 200 & 1&196.0&1 \\
 &BONN $\lambda=0.5$ & 199.7 & 0.06&181.6&0.26 \\
 &BONN $\lambda=1$ & 190.3 & 0.05&172.2&0.20 \\
 \hline
\multirow{2}{*}{$3\times3$-rooms }  &R-PG & -3.3 & 1&-14.9&1 \\
 &BONN $\lambda=1$ & -7.4 & 0.36&-16.3&0.61 \\ 
 \hline
\multirow{2}{*}{Lunar Lander}  &R-PG & 227.3 & 1&109.3&1 \\
 &BONN $\lambda=0.5$ & 221.2 & 0.16&91.6&0.07 \\
 &BONN $\lambda=5$ & 210.5 & 0.06&90.4&0.04 \\
\end{tabular}
}
\end{center}
\caption{Cost/reward values for the different environments, at different cost levels $\lambda$ and different stochasticity levels $\epsilon$.}\vspace{-0.3cm}
\label{table_results}
\end{table} 

Several environments were used to evaluate BONN in this setting: (i) \textbf{CartPole:} This is the classical cart-pole environment as implemented in the OpenAI Gym platform  \cite{gym} \footnote{https://gym.openai.com/} in which observations are $(position,angle,speed,angular speed)$, and possible actions are $right$ or $left$. The reward is $+1$ for every time step without failure. In this environment, the sizes of the networks used are $N_y=5$ and $N_{gru}=5$. (ii) \textbf{LunarLander:} This environment corresponds to the Lunar Lander environment proposed in OpenAI Gym where observations describe the position, velocity, angle of the agent and whether it is in contact with the ground or not, and possible actions are \textit{do nothing, fire left engine, fire main engine, fire right engine}. The reward is +100 if landing, +10 for each leg on the ground, -100 if crashing and -0.3 each time the main engine is fired. Here $N_y=10$ and $N_{gru}=10$. (iii) \textbf{$k \times k$-rooms:} This environment corresponds to a maze composed of $k \times k$ rooms with doors between them (see Figure \ref{fig/traj9}). The agent always starts at the upper-left corner, while the goal position is chosen randomly at each episode: it can be in any room, in any position and its position changes at each episode. The reward function is -1 when moving and +20 when reaching the goal, while 4 different actions are possible:  \textit{up,down,left} and \textit{right}. The observation describes the agent position, the position of the doors in the room, and the goal position if the goal is in the same room than the agent (i.e the agent only observes the current room). Note that this environment is much more difficult than other 4-rooms problems (introduced by \cite{sutton1999between}). In the latter, there is only one or two possible goal position(s) while in our case, the goal can be anywhere. Moreover, in our case, in a more realistic setting, the agent only observes the room it is in. Here $N_y=20$ and $N_{gru}=10$.

In simple environments like CartPole and LunarLander, the option model does not need to be recurrent and $o_t$ is just dependent on $x_t$ and $y_t$.

\paragraph{Results:} We compare BONN to a recurrent policy gradient (R-PG) with GRU cells. Note that R-PG has access to all observations ($x_t$ and $y_t$) at every time step and find the optimal policy in these environments, while BONN learns to use $y_t$ only from time to time.

We illustrate the quality of BONN in Table \ref{table_results}. There is two versions of each environment: a deterministic one ($\epsilon=0$) and a stochastic one ($\epsilon=0.25$) in which the movement of the agent can fail with probability $\epsilon$: in that case, a random transition is applied. In deterministic environments, we can see that the BONN model is able to perform as well as classical baselines while acquiring the observation only a few times per episode: for example in the Cartpole environment, the agent needs to use the observation only 6\% of the time. These results clearly show that in simple environments, there is no need to receive (observations) feedback permanently, and that a single planning step can generate several actions. However, in stochastic environments, observations are used more often and even then the performances degrade much more. Indeed, due to the stochasticity, it is not possible for the agent to deduce its position using only the chosen action (due to failure of actions, the agent will not know anymore in which state the environment is), and the observation thus needs to be acquired more often. It demonstrates the limits of open-loop policies in non predictable environments, justifying the use of a basic observation $x_t$ in the following.

%
\subsection{Using low-level and high-level observations}

\begin{figure}[t]
\includegraphics[width=0.95\linewidth]{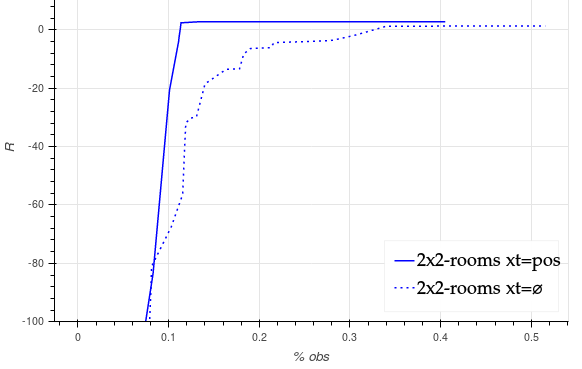}
\caption{Reward w.r.t. cost curves for $2\times 2$-rooms with a stochasticity level of $\epsilon=0.25$. }
\label{fig/compa}
\end{figure}

As seen above, the problem of open-loop control -- i.e control without any feedback from the environment -- is that the stochasticity of environment cannot be ``anticipated'' by the agent without receiving feedback. We study here a setting in which $x_t$ provides a simple and light information, while $y_t$ provides a richer information. The motivation is that $x_t$ can help to decide which action to take, without as much cognitive effort as when using the complete observation.

For that, we use another version of the $k\times k$-rooms environment where $x_t$ contains the agent position in the current room, while $y_t$ corresponds to the remaining information (i.e positions of the doors and position of the goal if it is in the room). The difference with the previous setting is that the agent has a permanent access to its position. In this version, we used $N_x=10$, $N_y=10$ and $N_{gru}=10$.

In a stochastic environment ($\epsilon=0.25$), BONN only uses $y_t$ 16\% of the time (versus 60\% in the blind setting) and even so achieve a cumulative reward of -18 (roughly the same).
We can see in Fig. \ref{fig/compa} the rewards w.r.t. cost curves obtained by computing the Pareto front over BONN models with different cost levels $\lambda$. We note that the drop of performance in $2\times 2$-rooms with $x_t=position$ happens at a lower cost than the one in $2\times2$-rooms environment with blind setting. Indeed, with $x_t=position$, the agent knows where it is at each time step and is more able to ``compensate'' the stochasticity of the environment than in the first case, in which the position is only available through $y_t$. Like in the deterministic blind setting, the agent is able to discover meaningful options and to acquire the relevant information only once by room -- see Section \ref{exp/analysis}.

This experiment shows the usefulness of permanently using a basic observation with a low cognitive cost, in contrast to temporary ``pure'' open-loop policies where no observation is used.

\subsection{Analysis of option discovered}
\label{exp/analysis}
\begin{figure}[t]
\begin{subfigure}[t]{0.25\textwidth}
\includegraphics[width=0.99\linewidth]{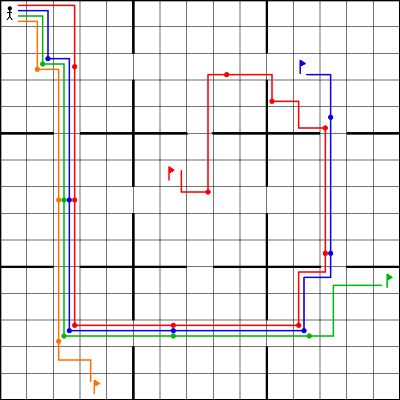}
\caption{}
\label{fig/traj9}
 \end{subfigure}
\begin{subfigure}[t]{0.20\textwidth} 
\includegraphics[width=0.98\linewidth]{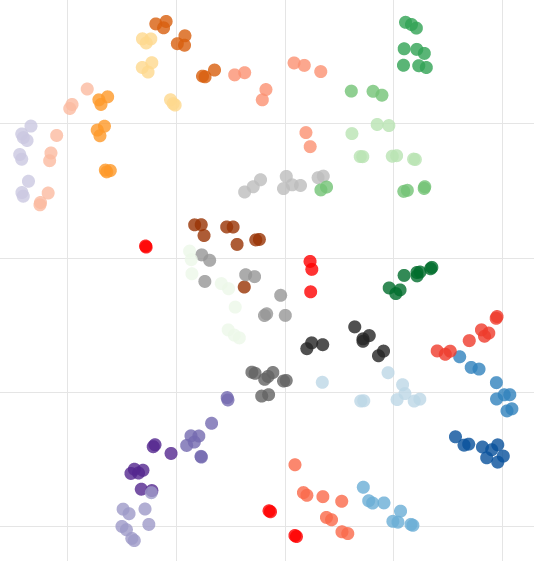}
\caption{}
\label{fig/tsne}
\end{subfigure}
\caption{(a) $3\times 3$-rooms: example of 4 trajectories (each one in different color) generated by the agent. Each point corresponds to a position where the agent decides to acquire $y_t$ and generates a new option: the agent uses only one option per room. (b) The options latent vectors visualized through the t-SNE algorithm. Similar colors mean the goal is in similar areas in the room, except for the red points that corresponds to the options used to reach one of the four possible doors, or when the goal is just near a door. }
\label{fig/traj}
\centering
\end{figure}

\begin{figure}[t]
\includegraphics[width=0.5\linewidth]{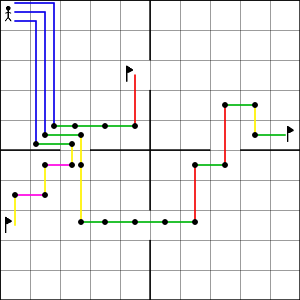}
\caption{$2\times 2$-rooms: Trajectories generated with the D-BONN model where $K=9$.}
\label{fig/traj_discret}
\end{figure}

Figures \ref{fig/traj9} illustrates trajectories generated by the agent in the $3\times3$-rooms environment, and the positions where the options are generated. We can see that the agent learns to observe $y_t$ only once in each room and that the agent uses the resulting option until it reaches another room. Thus the agent deducts from $y_t$ if he must move to another room, or reach the goal if it is in the current room. Note that the agent does not go directly to the goal room because it has no information about it. Seeing only the current room, it learns to explore the maze in a ``particular'' order until reaching the goal room. We have also visualized the options latent vectors using the t-SNE algorithm (Figure \ref{fig/tsne}). Similar colors (for example all green points) mean that the options computed correspond to observations for which the goals are in similar areas. We can for example see that all green options are close, showing that the latent option space effectively captures relevant information about options similarity. 

The D-BONN model has been experimented on the $2 \times 2$-rooms environment, and an example of generated trajectories is given in Figure \ref{fig/traj_discret}. Each color corresponds to one of the learned discrete options. One can see that the model is still able to learn a good policy, but the constraint over the fixed number of discrete options clearly decreases the quality of the obtained policy. It seems thus more interesting to use continuous options instead of discrete ones, the continuous options being regrouped in smooth clusters as illustrated in Figure \ref{fig/tsne}.

\subsection{Instructions as options}

We consider at last a setting in which $y_t$ is an information provided by an oracle,  while $x_t$ is a classical observation. The underlying idea is that the agent can choose an action based only on its observation, or use information from an optimal model with a higher cost. To study this case, we consider a maze environment where the maze is randomly generated for each episode (so the agent cannot memorize the exact map) and the goal and agent initial positions are also randomly chosen. The observation $x_t$ is the 9 cases surrounding the agent. The observation $y_t$ is a one hot vector of the action computed by a simple path planning algorithm that has access to the whole map. The parameters of the model are $N_x=10$ and $N_{gru}=5$ (no representation is used for $y_t$). Note that the computation of $y_t$ can be expensive, leading to the idea that it has a higher cognitive cost. 

\begin{figure}[t]
\begin{subfigure}[t]{0.235\textwidth}
\includegraphics[width=0.95\linewidth]{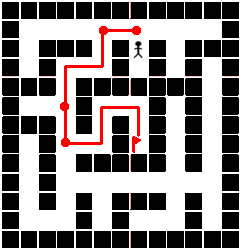}
\caption{}
\label{fig/maze_complete}
 \end{subfigure}
\begin{subfigure}[t]{0.235\textwidth} 
\includegraphics[width=0.96\linewidth]{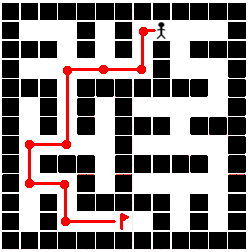}
\caption{}
\label{fig/maze_incomplete}
 \end{subfigure}
\caption{Examples of trajectories in the maze. Each point corresponds to a position where the agent decides to acquire $y_t$ and generates a new option: (a) Optimal learning with $y_t$ only used at each crossroad. (b) Incomplete learning where the agent also uses $y_t$ every time he must change direction. }\vspace{-0.3cm} 
\label{fig/maze}
\centering
\end{figure}

Two examples of generated trajectories are illustrated in Fig. \ref{fig/maze}. Figure \ref{fig/maze_incomplete} illustrates a generated trajectory \textbf{when learning is not finished}. It shows that, at a certain point, the low-controller has learned to follow a straight path but needs to ask for instructions at each cross-road, or when a change of direction is needed. Some relevant options have already emerged in the agent behavior but are not optimal. When \textbf{learning is finished} (Fig \ref{fig/maze_complete}), the agent now ask for instructions only at cross-roads where the decision is essential to reach the goal. Between crossroads, the agent learns to follow the corridors, which corresponds to an intuitive and realistic behavior. Note that future works will be done using outputs of more expensive models in lieu of a high-level observation $y_t$, and the BONN model seems to be an original way to study the \textit{Model Free/Model-based} paradigm proposed in neuroscience \cite{glascher2010states}.

\section{Conclusion and Perspectives}

We proposed a new model for learning options in POMDP in which the agent chooses when to acquire a more informative -- but costly -- observation at each time step. The model is learned in a budgeted learning setting in which the acquisition of the additional information, and thus the use of a new option, has a cost. The learned policy is a trade-off between the efficiency and the cognitive effort of the agent. In our setting, the options are handled through learned latent representations. Experimental results demonstrate the possibility of reducing the cognitive cost -- i.e. acquiring and computing information -- without a drastic drop in performances. We also show the benefit of using different levels of observation and the relevance of extracted options. This work opens different research directions. One is to study if BONN can be applied in multi-task reinforcement learning problems (the environment $k\times k$-rooms,  since the goal position is randomly chosen at each episode, can be seen as a  multi-task problem). Another question would be to study problems where many different observations can be acquired by the agent at different costs - e.g, many different sensors on a robot. Finally, a promising perspective is learning how and when to interact with another expensive model.

 \section*{Acknowlegments}

 This work has been supported within the Labex SMART supported by French state funds managed by the ANR within the Investissements d’Avenir programme under reference ANR-11-LABX-65.

\newpage

\small
\bibliographystyle{named}
\bibliography{ijcai17}

\begin{thebibliography}{}

\bibitem[\protect\citeauthoryear{Bacon and Precup}{2015a}]{bacondeliberate}
Pierre-Luc Bacon and Doina Precup.
\newblock Learning with options: Just deliberate and relax.
\newblock 2015.

\bibitem[\protect\citeauthoryear{Bacon and Precup}{2015b}]{bacon2015option}
Pierre-Luc Bacon and Doina Precup.
\newblock The option-critic architecture.
\newblock In {\em NIPS Deep Reinforcement Learning Workshop}, 2015.

\bibitem[\protect\citeauthoryear{Botvinick \bgroup \em et al.\egroup
  }{2009}]{hie}
Matthew Botvinick, Yael Niv, and Andrew~C. Barto.
\newblock Hierarchically organized behavior and its neural foundations: A
  reinforcement-learning perspective.
\newblock {\em cognition}, 113.3, 2009.

\bibitem[\protect\citeauthoryear{Brockman \bgroup \em et al.\egroup
  }{2016}]{gym}
Greg Brockman, Vicki Cheung, Ludwig Pettersson, Jonas Schneider, John Schulman,
  Jie Tang, and Wojciech Zaremba.
\newblock Openai gym, 2016.

\bibitem[\protect\citeauthoryear{Bui \bgroup \em et al.\egroup }{2002}]{ahmm}
Hung~Hai Bui, Svetha Venkatesh, and Geoff West.
\newblock Policy recognition in the abstract hidden markov model.
\newblock {\em Journal of Artificial Intelligence Research}, 17:451--499, 2002.

\bibitem[\protect\citeauthoryear{Cho \bgroup \em et al.\egroup
  }{2014}]{cho2014learning}
Kyunghyun Cho, Bart Van~Merri{\"e}nboer, Caglar Gulcehre, Dzmitry Bahdanau,
  Fethi Bougares, Holger Schwenk, and Yoshua Bengio.
\newblock Learning phrase representations using rnn encoder-decoder for
  statistical machine translation.
\newblock {\em arXiv preprint arXiv:1406.1078}, 2014.

\bibitem[\protect\citeauthoryear{Chung \bgroup \em et al.\egroup
  }{2016}]{chung2016hierarchical}
Junyoung Chung, Sungjin Ahn, and Yoshua Bengio.
\newblock Hierarchical multiscale recurrent neural networks.
\newblock {\em arXiv preprint arXiv:1609.01704}, 2016.

\bibitem[\protect\citeauthoryear{Contardo \bgroup \em et al.\egroup
  }{2016}]{contardo2016recurrent}
Gabriella Contardo, Ludovic Denoyer, and Thierry Arti{\`e}res.
\newblock Recurrent neural networks for adaptive feature acquisition.
\newblock In {\em International Conference on Neural Information Processing},
  pages 591--599. Springer International Publishing, 2016.

\bibitem[\protect\citeauthoryear{Daniel \bgroup \em et al.\egroup
  }{2016}]{danielprobabilistic}
Christian Daniel, Herke van Hoof, Jan Peters, and Gerhard Neumann.
\newblock Probabilistic inference for determining options in reinforcement
  learning.
\newblock 2016.

\bibitem[\protect\citeauthoryear{Dayan and Hinton}{1993}]{dayan1993feudal}
Peter Dayan and Geoffrey~E Hinton.
\newblock Feudal reinforcement learning.
\newblock In {\em Advances in neural information processing systems}, pages
  271--271. Morgan Kaufmann Publishers, 1993.

\bibitem[\protect\citeauthoryear{Dezfouli and
  Balleine}{2012}]{dezfouli2012habits}
Amir Dezfouli and Bernard~W Balleine.
\newblock Habits, action sequences and reinforcement learning.
\newblock {\em European Journal of Neuroscience}, 35(7):1036--1051, 2012.

\bibitem[\protect\citeauthoryear{Dezfouli and
  Balleine}{2013}]{dezfouli2013actions}
Amir Dezfouli and Bernard~W Balleine.
\newblock Actions, action sequences and habits: evidence that goal-directed and
  habitual action control are hierarchically organized.
\newblock {\em PLoS Comput Biol}, 9(12):e1003364, 2013.

\bibitem[\protect\citeauthoryear{Dietterich}{1998}]{dietterich1998maxq}
Thomas~G Dietterich.
\newblock The maxq method for hierarchical reinforcement learning.
\newblock In {\em ICML}, pages 118--126. Citeseer, 1998.

\bibitem[\protect\citeauthoryear{Dulac-Arnold \bgroup \em et al.\egroup
  }{2012}]{dulac2012sequential}
Gabriel Dulac-Arnold, Ludovic Denoyer, Philippe Preux, and Patrick Gallinari.
\newblock Sequential approaches for learning datum-wise sparse representations.
\newblock {\em Machine Learning}, 89(1-2):87--122, 2012.

\bibitem[\protect\citeauthoryear{Florensa \bgroup \em et al.\egroup
  }{2016}]{florensa2016stochastic}
Carlos Florensa, Yan Duan, and Pieter Abbeel.
\newblock Stochastic neural networks for hierarchical reinforcement learning.
\newblock {\em ICLR}, 2016.

\bibitem[\protect\citeauthoryear{Gl{\"a}scher \bgroup \em et al.\egroup
  }{2010}]{glascher2010states}
Jan Gl{\"a}scher, Nathaniel Daw, Peter Dayan, and John~P O'Doherty.
\newblock States versus rewards: dissociable neural prediction error signals
  underlying model-based and model-free reinforcement learning.
\newblock {\em Neuron}, 66(4):585--595, 2010.

\bibitem[\protect\citeauthoryear{Gregor \bgroup \em et al.\egroup
  }{2016}]{gregor2016variational}
Karol Gregor, Danilo~Jimenez Rezende, and Daan Wierstra.
\newblock Variational intrinsic control.
\newblock {\em arXiv preprint arXiv:1611.07507}, 2016.

\bibitem[\protect\citeauthoryear{Hansen \bgroup \em et al.\egroup
  }{1996}]{hansen1996reinforcement}
Eric~A Hansen, Andrew~G Barto, and Shlomo Zilberstein.
\newblock Reinforcement learning for mixed open-loop and closed-loop control.
\newblock In {\em NIPS}, pages 1026--1032, 1996.

\bibitem[\protect\citeauthoryear{Hauskrecht \bgroup \em et al.\egroup
  }{1998}]{hauskrecht1998hierarchical}
Milos Hauskrecht, Nicolas Meuleau, Leslie~Pack Kaelbling, Thomas Dean, and
  Craig Boutilier.
\newblock Hierarchical solution of markov decision processes using
  macro-actions.
\newblock In {\em Proceedings of the Fourteenth conference on Uncertainty in
  artificial intelligence}, pages 220--229. Morgan Kaufmann Publishers Inc.,
  1998.

\bibitem[\protect\citeauthoryear{Heess \bgroup \em et al.\egroup
  }{2016}]{heess2016learning}
Nicolas Heess, Greg Wayne, Yuval Tassa, Timothy Lillicrap, Martin Riedmiller,
  and David Silver.
\newblock Learning and transfer of modulated locomotor controllers.
\newblock {\em arXiv preprint arXiv:1610.05182}, 2016.

\bibitem[\protect\citeauthoryear{Keramati \bgroup \em et al.\egroup
  }{2011}]{keramati2011speed}
Mehdi Keramati, Amir Dezfouli, and Payam Piray.
\newblock Speed/accuracy trade-off between the habitual and the goal-directed
  processes.
\newblock {\em PLoS Comput Biol}, 7(5):e1002055, 2011.

\bibitem[\protect\citeauthoryear{Kingma and Ba}{2014}]{kingma2014adam}
Diederik Kingma and Jimmy Ba.
\newblock Adam: A method for stochastic optimization.
\newblock {\em arXiv preprint arXiv:1412.6980}, 2014.

\bibitem[\protect\citeauthoryear{Kool and Botvinick}{2014}]{kool2014labor}
Wouter Kool and Matthew Botvinick.
\newblock A labor/leisure tradeoff in cognitive control.
\newblock {\em Journal of Experimental Psychology: General}, 143(1):131, 2014.

\bibitem[\protect\citeauthoryear{Kulkarni \bgroup \em et al.\egroup
  }{2016}]{kulkarni2016hierarchical}
Tejas~D Kulkarni, Karthik~R Narasimhan, Ardavan Saeedi, and Joshua~B Tenenbaum.
\newblock Hierarchical deep reinforcement learning: Integrating temporal
  abstraction and intrinsic motivation.
\newblock {\em arXiv preprint arXiv:1604.06057}, 2016.

\bibitem[\protect\citeauthoryear{Mnih \bgroup \em et al.\egroup
  }{2016}]{mnih2016strategic}
Volodymyr Mnih, John Agapiou, Simon Osindero, Alex Graves, Oriol Vinyals, Koray
  Kavukcuoglu, et~al.
\newblock Strategic attentive writer for learning macro-actions.
\newblock {\em arXiv preprint arXiv:1606.04695}, 2016.

\bibitem[\protect\citeauthoryear{Parr and
  Russell}{1998}]{parr1998reinforcement}
Ronald Parr and Stuart Russell.
\newblock Reinforcement learning with hierarchies of machines.
\newblock {\em Advances in neural information processing systems}, pages
  1043--1049, 1998.

\bibitem[\protect\citeauthoryear{Sutton \bgroup \em et al.\egroup
  }{1999}]{sutton1999between}
Richard~S Sutton, Doina Precup, and Satinder Singh.
\newblock Between mdps and semi-mdps: A framework for temporal abstraction in
  reinforcement learning.
\newblock {\em Artificial intelligence}, 112(1):181--211, 1999.

\bibitem[\protect\citeauthoryear{Wierstra \bgroup \em et al.\egroup
  }{2010}]{rpg}
Daan Wierstra, Alexander F{\"{o}}rster, Jan Peters, and J{\"{u}}rgen
  Schmidhuber.
\newblock Recurrent policy gradients.
\newblock {\em Logic Journal of the {IGPL}}, 18(5):620--634, 2010.

\end{thebibliography}

\end{document}